\documentclass[conference]{IEEEtran}
\IEEEoverridecommandlockouts
\usepackage{cite}
\usepackage{amsmath,amssymb,amsfonts}
\usepackage{algorithmic}
\usepackage{graphicx}
\usepackage{textcomp}
\usepackage{xcolor}
\usepackage{algorithmic}
\usepackage{algorithm}

\def\BibTeX{{\rm B\kern-.05em{\sc i\kern-.025em b}\kern-.08em
    T\kern-.1667em\lower.7ex\hbox{E}\kern-.125emX}}
\begin{document}

\title{Enhancing the Robustness of QMIX against State-adversarial Attacks\\
}
\author{\IEEEauthorblockN{1\textsuperscript{st} Weiran Guo}
\IEEEauthorblockA{\textit{Department of Computer Sciences} \\
\textit{Tongji University}\\
Shanghai, China \\
azureeeeeguo@gmail.com}
\and
\IEEEauthorblockN{2\textsuperscript{nd} Guanjun Liu}
\IEEEauthorblockA{\textit{Department of Computer Sciences} \\
\textit{Tongji University}\\
Shanghai, China \\
liuguanjun@tongji.edu.cn}
\and
\IEEEauthorblockN{3\textsuperscript{rd} Ziyuan Zhou}
\IEEEauthorblockA{\textit{Department of Computer Sciences} \\
\textit{Tongji University}\\
Shanghai, China \\
ziyuanzhou@tongji.edu.cn}
\and
\IEEEauthorblockN{4\textsuperscript{th} Ling Wang}
\IEEEauthorblockA{\textit{College of Transportation Engineering} \\
\textit{Tongji University}\\
Shanghai, China \\
wang\_ling@tongji.edu.cn}
\and
\IEEEauthorblockN{5\textsuperscript{th} Jiacun Wang}
\IEEEauthorblockA{\textit{Department of Computer Science and Software Engineering} \\
\textit{Monmouth University}\\
West Long Branch, USA \\
jwang@monmouth.edu}
}

\maketitle

\begin{abstract}
Deep reinforcement learning (DRL) performance is generally impacted by state-adversarial attacks, a perturbation applied to an agent's observation. Most recent research has concentrated on robust single-agent reinforcement learning (SARL) algorithms against state-adversarial attacks. Still, there has yet to be much work on robust multi-agent reinforcement learning. Using QMIX, one of the popular cooperative multi-agent reinforcement algorithms, as an example, we discuss four techniques to improve the robustness of SARL algorithms and extend them to multi-agent scenarios. To increase the robustness of multi-agent reinforcement learning (MARL) algorithms, we train models using a variety of attacks in this research. We then test the models taught using the other attacks by subjecting them to the corresponding attacks throughout the training phase. In this way, we organize and summarize techniques for enhancing robustness when used with MARL.
\end{abstract}


\section{Introduction}
When it comes to completing activities that call for reaping the best rewards, DRL has made significant progress. However, many studies indicate that many agents who have received DRL training are still vulnerable to attacks during the testing phase. The state-adversarial attack is one of the assaults; it induces the agent to observe without altering the current environment. In practical application, it refers to the discrepancy between an ideal setting and the real world. For instance, if minor disturbances brought on by sensor limitations or malicious attacks are added to the state input in an autonomous driving task, they may lead to the car making judgments against expectations and leading to negative outcomes. Thus, strengthening the DRL algorithm's robustness is highly crucial.

Several studies suggest that there are techniques to increase the robustness of SARL, but there has yet to be much discussion on improving the robustness of MARL. The latter, though, is a subject worth exploring. A multi-agent environment is a common condition observed in various industries, from games to transportation, which is one of the key reasons. Moreover, most multi-agent problems are trickier to solve than single-agent ones. The agents are not distinct individuals but rather are connected. The total return of the entire system may drop if one of the agents is assaulted. 

The above factors encourage us to research multi-agent reinforcement learning algorithms resistant to state-adversarial perturbations. The following is this paper's significant contributions:
\begin{itemize}
\item We list four training strategies: \textit{gradient-based adversary}, \textit{policy regularization}, \textit{alternating training with learned adversaries} (ATLA) and \textit{policy adversarial actor director} (PA-AD), all of which have been used to date to adapt SARL algorithms to the multi-agent scenario.
\item The cooperative multi-agent reinforcement learning (c-MARL) algorithm QMIX\cite{b1} is used as an example, and four alternative ways are used to improve its robustness in a MARL environment. We next attack the multi-agent body system that was trained using these methods. 
\item As a result of the experiments, conclusions are reached, and a comparison of these four methodologies reveals their benefits and drawbacks.
\end{itemize}

The paper is structured as follows. Section II sorts out related works about enhancing the robustness of both SARL and MARL. Section III introduces the background of the problem we are discussing. Section IV describes four existing methods applied to SARL, transferring them to MARL. Section V shows the results of the experiments performed. Finally, conclusions and analysis are offered.

\section{Related Work}
\subsection{Adversarial Attack and Adversarial Training in SARL}
There are mainly three types of adversarial attack and training methods to bolster the robustness of SARL. 1) Modifying the loss function. Zhang et al.\cite{b2} alter the loss function of the training stage with regularization, making it more consistent with the latent mathematical relations of the reinforcement learning problem. Oikarinen et al.\cite{b3} propose RADIAL-RL to derive the adversarial loss. 2) Applying heuristic attacks. Pattanaik et al.\cite{b4} use attacks for machine learning image recognition, e.g., FGSM\cite{b5}, PGD\cite{b6}, etc., on the state observation of the agent. 3) Training a network for the adversary. Tretschk et al.\cite{b7} attack the agent sequentially using the most current adversarial attack method, Adversarial Transformer Network (ATN)\cite{b8}, which learns to create the assault and is simple to integrate into the policy network. Zhang et al.\cite{b9} propose ATLA, and Sun et al.\cite{b10} propose PA-AD, both utilizing an opponent that generates the optimal adversary to teach the agent to be resilient to various attack strengths.
\subsection{Adversarial Attack in MARL}
Attacks on multi-agent systems and evaluation frameworks have been a significant focus of much of the existing research on MARL robustness. Guo et al.\cite{b11} propose MARLSafe, a robustness testing framework for c-MARL algorithms that evaluates three aspects of attacks, including the one involving state observation. Pham et al.\cite{b12} propose the first model-based adversarial attack framework for c-MARL. Lin et al.\cite{b13} and Hu and Zhang\cite{b14}, respectively, choose to apply attacks on one of the agents in the multi-agent system and during a few of the timesteps, proving that even if the agents are not all attacked the whole time, they still perform poorly. 

\subsection{Adversarial Training in MARL}
Zhang et al.\cite{b15} propose robust Markov games that consider model uncertainty and improve model performance using function approximation and mini-batch updates. To overcome the difficulty of training resilient policies under adversarial state perturbations based on a gradient descent ascent technique, Han et al.\cite{b16} offer the Robust Multi-Agent Adversarial Actor-Critic (RMA3C) algorithm. Zhou and Liu\cite{b17} propose a brand-new objective function and a repetitive regularization method to enhance MARL's defending ability. Shi et al.\cite{b18} consider generalizability and use random noise to bridge the real and virtual settings.
\section{Background}
\subsection{c-MARL Algorithm: QMIX}
QMIX is a value-based method for centralized, end-to-end training and decentralized execution. QMIX uses a network that, using just local data, estimates the joint action values as a complicated nonlinear combination of each agent's values. To ensure consistency between the centralized and decentralized methods, a constraint is placed on the structure. The joint action-value function is monotonic concerning each agent's action-value function.

\begin{figure}[h]
\centerline{\includegraphics[width=0.40\textwidth]{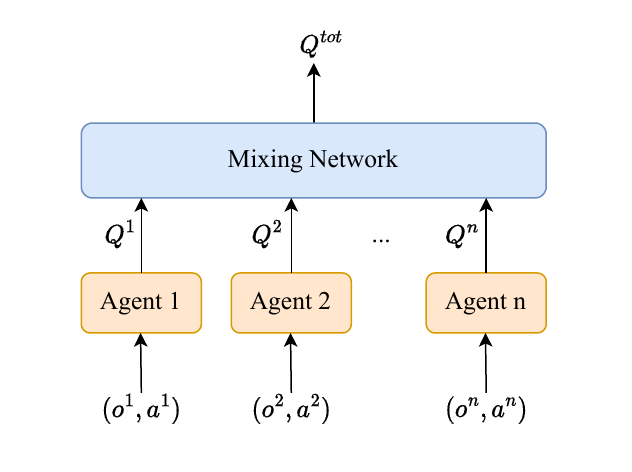}}
\caption{The macro architecture of QMIX.}
\label{fig1}
\end{figure}

Fig.~\ref{fig1} illustrates a macroscopic structure of the QMIX network. The action-value function part for single agents and a joint action-value function part for several agents comprise the two primary components of the QMIX network architecture. A single-agent action-value function network receives the corresponding agent's partial observations $o^{i}$ and actions $a^{i}$ as inputs, generating the action-value function of the agent. The joint action-value function network of QMIX accepts the $Q$-values of all the agents' selected actions. This mixing network then calculates the joint $Q$-value $Q^{tot}$ with hyper networks generating the necessary weights and biases.

\subsection{State-adversarial Stochastic Game}
A stochastic game is a game process that involves one or more players and state probability transitions, a process in which the players' actions gain rewards and cause the occurrence of the subsequent stochastic state. 
The state-adversarial stochastic game fits with the multi-agent case under perturbation we discuss in this paper. A state-adversarial stochastic game (SaSG) is defined in \cite{b13} with state perturbations involved in the stochastic game. Define that the state adversary only changes with the current state and is unaffected by time. 

\subsection{Dec-POMDP with State Adversary}
In most scenarios, the agents can only be described from the observed part of the environment and therefore are described using a decentralized partially observable Markov decision process (Dec-POMDP)\cite{b19}. With perturbations involved, we use a state-adversarial Dec-POMDP to model the subject we are studying. It can be defined as a tuple $<S, \{A^{i}\}^{N}, P, \{R^{i}\}^{N}, \{O^{i}\}^{N}, \{B^{i}\}^{M},\gamma>$, where $N$ is the number of agents, $M$ is the number of the attacked agents, $A^{i}$ is the action space of agent $i$ with $A$ the set of joint actions, $P$ is the state transition probabilities which refers to the probability distributions based on the current state and actions, $R^{i}$ is the reward function of agent $i$ with $R$ the global reward function of the multi-agent system, $O^{i}$ is the observation space of agent $i$ with $O$ the joint observations of all agents, $B^{i}$ is the set of adversarial states applied on agent $i$, and $\gamma\in[0,1]$ serves as the discount factor. 

Dec-POMDP in c-MARL is a Markov decision process in which a group of agents cooperate to maximize their total team reward while only obtaining local information. To be more specific, each agent $i(0<i<N)$ in the team receives partial observation $o_{t}^{i}$ as its input at timestep $t$ when there are $N$ agents participating in the cooperating task. Each agent $i$ performs an action $a_{t}^{i}$ according to the observation it collects, trying to get closer to the maximal total team reward $R$.

The observation inputs that the agents receive in reality and those that they are meant to receive differ when state perturbations are present in the c-MARL environment. Because the observations are perturbed, the observed state that the agent takes in changes from $o^{i}$ to $\Tilde{o}^{i}$. According to the observation, agent $i$'s output actions vary. The chosen action may change if the observation does. The agent's Q-value changes from $Q^{i}(o^{i},a^{i})$ to $Q^{i}(\Tilde{o}^{i},a^{i})$ as a result of such a modification, changing the joint value $Q^{tot}$ as well. The c-MARL algorithm operates under the assumption that $Q^{tot}$ that must be obtained represents the current optimal solution and that a change in $Q^{tot}$ results in a reduction in the overall benefit.

\section{Methods}
In deep learning, training employs adversarial assaults to optimize the multi-agent system's worst-case reward to resolve the max-min problem. This increases the robustness of the model. This concept is also applied in MARL, where the effect of adversarial training is measured, and the max-min problem is solved. It is known that in QMIX, a monotonic constraint relationship is satisfied between the individual $Q^{i}$ and the global $Q^{tot}$, so we derive the objective function of the expected reward function in terms of the attacked single agents in our study of the problem:
 \begin{equation}
\mathop{\max}_{\pi^{i}}\mathop{\min}_{\Tilde{o}^{i}}\mathop{\sum}_{a^{i}}\pi^i(a_{i}|\Tilde{o}^{i}){Q_{\varphi^i}}(\Tilde{o}^{i},a^{i})
\end{equation}
where $\pi^{i}$ is the policy of agent $i$, $\Tilde{o}^{i}$is the perturbed state observation received by agent $i$, and $\varphi^i$ is the parameters of agent $i$'s RNN network. The inner minimum is crucial because we need to identify a worst-case situation; hence, we must choose an optimal adversary.

Aiming to minimize the multi-agent system's overall reward is important to choose the best opponent to impose on the observations. This is the main idea behind the attacking policy on observations during training and testing, with the objective function:
\begin{equation}
\underset{\Tilde{\pi}}{\operatorname{argmin}} \sum_{i=0}^\infty\gamma^{t}R(s_t,a_t^1,...a_t^M,s_{t+1})
\end{equation}
where $\Tilde{\pi}:=(\Tilde{\pi}^{1},...\Tilde{\pi}^{M})$ is the set of victim agents' policies. The goal of the objective function is to make the total reward $R$ as low as possible.

\subsection{Gradient-based Adversary in MARL}
\begin{figure}[tb]
\centerline{\includegraphics[width=0.40\textwidth]{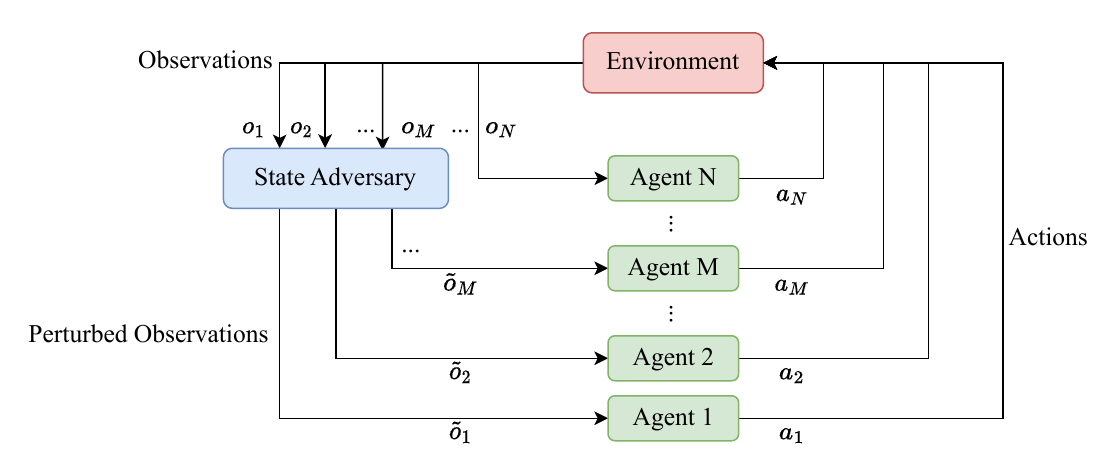}}
\caption{Gradient-based adversarial attacks in MARL. The state adversary uses a gradient-based optimization method.}
\label{fig2}
\end{figure}

Gradient-based adversarial examples are proposed in image classification issues. In an MARL case, the goal of the gradient-based adversarial approach is to reduce the total reward of the multi-agent system by generating adversarial samples and attacking $M$ agents in the training phase, allowing QMIX to maximize the overall reward after the reduction as much as possible. Fig.~\ref{fig2} illustrates how a gradient-based adversary is used in a MARL instance.

A gradient-based adversary crafts the optimal max-norm constrained perturbation based on the chosen actions and the ones with the highest probability, which provide the most rewards in QMIX. Therefore we can define the loss function as the cross-entropy loss $L(\theta^i,\Tilde{o}^i,u^i)$ where $u^i$ is the target action with the highest probability of agent $i$ and $\theta^i$ is the parameters of the gradient-based adversary that attacks agent $i$.

Popular techniques for producing adversarial examples include FGSM, PGD, etc. We use FGSM when training robust QMIX, with $\Tilde{o}^i=o^i+\delta^i$ where $\delta^i$ is the perturbation imposed on agent $i$. The perturbation can be described as:
\begin{equation}
\delta^i=\epsilon sign(\nabla_{\Tilde{o}^i}L(\theta^i,\Tilde{o}^i,u^i))
\end{equation}
where $\epsilon$ is the $l_{\infty}$ norm perturbation budget. FGSM aims to produce perturbations that attempt to diverge the chosen action from the action with the highest gain; however, because these perturbations lack a target, the agents being attacked may not always act in the direction of the action with the lowest reward. In \cite{b10}, it is proven that gradient-based techniques have the possibility of not generating optimal attacks.

\subsection{Policy Regularization in MARL}
The policy regularization method uses a robust policy hinge-like regularizer as a defensive strategy to keep the best action unchanged after disturbance. \cite{b10} shows that the difference between the pre-perturbation value function and the post-perturbation value function in a multi-agent environment may be limited to a range if the disparities in the action distributions of the agents are not too significant. 

As for any of the agents in the QMIX algorithm, it chooses the one with the largest $Q$-value when picking an action, therefore $\pi(a^i|o^i)=1$ if $a$ is the best action and 0 otherwise. We use the total variation distance to measure the discrepancy between the action distributions: 
\begin{equation}
\begin{aligned}
&D_{TV}(\pi(\cdot|o^i),\pi(\cdot|\Tilde{o}^i)) =\\
&[\mathop{argmax}_{a^i}\pi(a^i|o^i)\neq\mathop{argmax}_{a^i}\pi(a^i|\Tilde{o}^i)]
\end{aligned}
\end{equation}
where $[\mathop{argmax}_{a^i}\pi(a^i|o^i)\neq\mathop{argmax}_{a^i}\pi(a^i|\Tilde{o}^i)]$ is the Iverson bracket, the value of which is 1 if the statement inside is true and 0 otherwise. The total variance decreases if the action chosen in the perturbed state is the same as the optimum action in the clean state.

We add a regular term to the loss function that minimizes the difference in action distributions, indirectly reducing the interference with the value function. In QMIX, the loss function calculates the loss of the whole multi-agent system and therefore calculates the overall regular term:
\begin{equation}
\begin{aligned}
&\mathcal{L}_{reg}(\pmb \varphi):= \\
&\mathop{\sum_i}\max\{\mathop{\sum_{o^i}}\mathop{\max_{\Tilde{o}^i\in B^i}\mathop{\max_{a^i\neq a^i_*}}}Q_{\varphi^i}(\Tilde{o}^i,a^i)-Q_{\varphi^i}(\Tilde{o}^i,a_*^i),-c\}
\end{aligned}
\end{equation}
where $\pmb \varphi:=(\varphi^1,..., \varphi^N)$ is the set of the parameters of all agents' RNN networks, $a_*^i$ is the action that reaches the largest value function of agent $i$, and $c$ is a positive constant to constrain the regular term.
\begin{equation}
\mathcal{L}_{tot}(\pmb \varphi)=\mathcal{L}(\pmb \varphi)+\kappa\mathcal{L}_{reg}(\pmb \varphi)
\end{equation}
where $\kappa$ is the weighting factor that balances the two components. This equation is used to set a limit on the QMIX's TD error as well as the value functions' perturbation.

\subsection{ATLA in MARL}
\begin{figure}[tb]
\centerline{\includegraphics[width=0.40\textwidth]{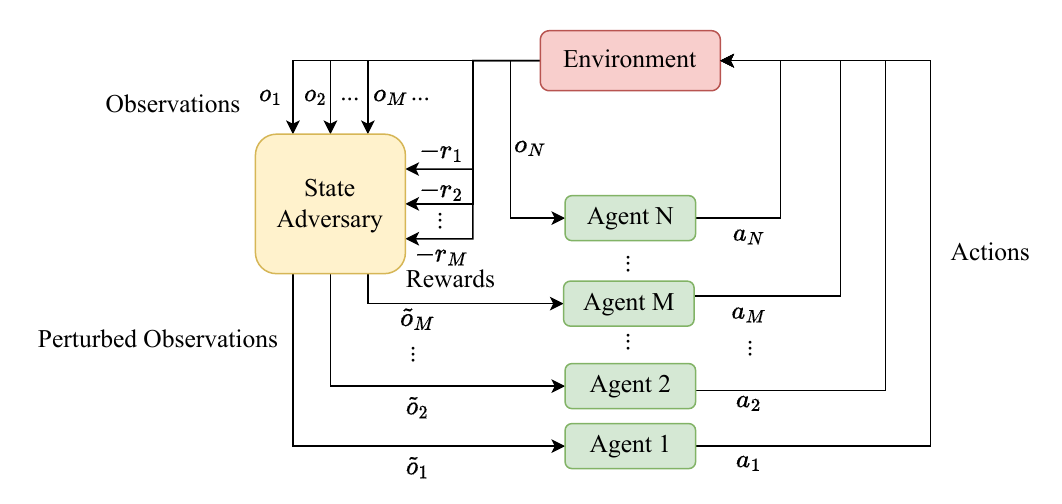}}
\caption{ATLA attacks in MARL. The state adversary is a neural network.}
\label{fig3}
\end{figure}

According to \cite{b17}, the joint optimal adversarial perturbation exists and is unique. The idea of ATLA in MARL is to train a network to output the best perturbation states and add them to the observations of the agents. The objective function of this network aims to minimize the total reward of the multi-agent system—the attack process shown in Fig.~\ref{fig3}.

The problem can be constructed as a stochastic game and solved by multi-agent reinforcement learning. As mentioned above, a multi-agent algorithm that can output continuous actions, such as MAPPO or MADDPG, can be used to update such a network. To make the rewards of the victim agents as little as possible, the idea is similar to building another multi-agent team in which the action space of each agent's output is a state perturbation and the reward earned is the opposite number of the reward of the attacked multi-agent group. The training process of the adversary denoted by $f$ can be seen in Algorithm~\ref{alg:atla}.

\begin{algorithm}[!h]
    \caption{Multi-agent ATLA}
    \label{alg:atla}
    \renewcommand{\algorithmicrequire}{\textbf{Input:}}
    \renewcommand{\algorithmicensure}{\textbf{Output:}}
    \begin{algorithmic}[1]
        \REQUIRE Initialization of adversary's policy $f$; victim policy $\pi$; start state $s_0$; start observation $o_0$   
        \FOR{$t=0,1,2,...$}
            \STATE Adversary samples the perturbing observations $\Tilde{\textbf{o}}_t\sim f(\cdot|\textbf{o}_t)$ 
            \FOR{agent $i=1\ to\ N$}
                \STATE Victim agent $i$ takes action $a^i_t\sim \pi(\cdot|\Tilde{o}^i_t)$, receives $r^i_t$ and $o^i_{t+1}$
                \STATE $\textbf{r}_t=\textbf{r}_t\cup\{r^i_t\}$
                \STATE $\textbf{o}_{t+1}=\textbf{o}_{t+1}\cup\{o^i_{t+1}\}$
            \ENDFOR
            \STATE Multi-agent system proceeds to $s_{t+1}$
            \STATE Adversary saves $(s_t,\textbf{o}_t,\hat{\textbf{a}}_t,-\textbf{r}_t,s_{t+1},\textbf{o}_{t+1})$ to its buffer
            \STATE Adversary updates policy $v$
            \ENDFOR
    \end{algorithmic}
\end{algorithm}

To train QMIX, we use the MAPPO algorithm as the adversary. For each agent $i$, the adversary generates a perturbation $\delta^i$ and adds it to $o^i$ with the original observations, perturbations, negative rewards, and the following observations stored in its replay buffer. Then it is trained to produce optimal attacks to minimize the total reward of the whole multi-agent system. This is an excellent way to construct a worst-case scenario, equivalent to two groups of multiple agents cross-training against each other.

\subsection{PA-AD in MARL}
\begin{figure}[tb]
\centerline{\includegraphics[width=0.40\textwidth]{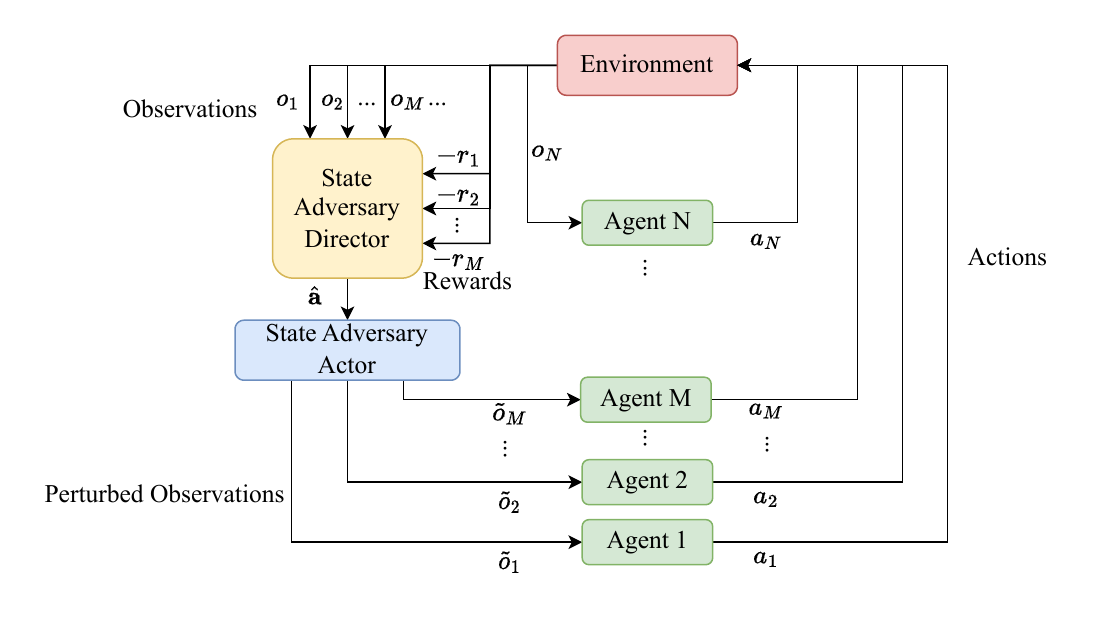}}
\caption{PA-AD attacks in MARL. The state adversary actor is a neural network that generates the policy-perturbing direction $\hat{\textbf{a}}$. The state adversary actor produces an optimal attack according to $\hat{\textbf{a}}$.}
\label{fig4}
\end{figure}

\cite{b10} demonstrates how perturbations to policies are analogous to evasive and action attacks, which is the underlying idea of PA-AD. Therefore, we can transfer the perturbations on states to the attacks on the agents' policies. The fundamental methods of PA-AD and ATLA are comparable; both train a neural network to manufacture an adversary. However, PA-AD uses a director and an actor to carry out the attacking objective instead of producing state perturbations directly. The director denoted by $v$ points out the optimal adversarial direction of the perturbation while the actor acts toward this direction using gradient-based adversaries. Using an RL algorithm (e.g., MAPPO, QMIX), the director resolves an RL problem where its actions are the adversarial directions and rewards are the negative values of the agents' rewards. The actor denoted by $g$ deals with the optimization problem, finding the optimal adversaries according to the directions the director points at with a supervised learning optimization method (e.g., FGSM, PGD). How the director and the actor perform is shown in Fig.~\ref{fig4}. The training process of the director can be seen in Algorithm~\ref{alg:pa-ad}.

\begin{algorithm}[!h]
    \caption{Multi-agent PA-AD}
    \label{alg:pa-ad}
    \renewcommand{\algorithmicrequire}{\textbf{Input:}}
    \begin{algorithmic}[1]
        \REQUIRE Initialization of director's policy $v$; victim policy $\pi$; start state $s_0$; start observation $o_0$   
        \FOR{$t=0,1,2,...$}
            \STATE Director samples the perturbing directions $\hat{\textbf{a}}_t\sim v(\cdot|\textbf{o}_t)$
            \STATE Actor perturbs $\textbf{o}_t$ to $\Tilde{\textbf{o}}_t=g(\hat{\textbf{a}}_t,\textbf{o}_t)$  
            \FOR{agent $i=1\ to\ N$}
                \STATE Victim agent $i$ takes action $a^i_t\sim \pi(\cdot|\Tilde{o}^i_t)$, receives $r^i_t$ and $o^i_{t+1}$
                \STATE $\textbf{r}_t=\textbf{r}_t\cup\{r^i_t\}$
                \STATE $\textbf{o}_{t+1}=\textbf{o}_{t+1}\cup\{o^i_{t+1}\}$
            \ENDFOR
            \STATE Multi-agent system proceeds to $s_{t+1}$    
            \STATE Director saves $(s_t,\textbf{o}_t,\hat{\textbf{a}}_t,-\textbf{r}_t,s_{t+1},\textbf{o}_{t+1})$ to its buffer
            \STATE Director updates policy $v$
            \ENDFOR
    \end{algorithmic}
\end{algorithm}

Another QMIX network can be used directly to train QMIX. To reduce the overall returns of the multi-agent system, its replay buffer receives negative values of the rewards from the multi-agent system. The perturbing directions are $\hat{\textbf{a}}\sim v(\cdot|s_t)$ where $v$ is the director's policy. Then use FGSM to put the actions generated by the adversarial QMIX into consideration, with the loss function as: 
\begin{equation}
L_{tar}(\pmb\phi^i,\Tilde{o}^i,u^i,\hat{a}^i)=-L_1(\phi_1^i,\Tilde{o}^i,\hat{a}^i)+L_2(\phi_2^i,\Tilde{o}^i,u^i)
\end{equation}
where $\pmb\phi^i=(\phi_1^i,\phi_2^i)$ is the set of the parameters of the two loss functions,  $\hat{a}^i$ is the action of adversarial QMIX and also the perturbation direction placed on agent $i$. Both $L_1$ and $L_2$ can be entropy-loss functions.

Theoretically, PA-AD overcomes the drawbacks of ATLA and gradient-based approaches, addressing the issue of computational complexity brought on by the large state space and offering the best guidance for optimization approaches that would otherwise lack objectives.

\section{Experiments}
We put the four adversarial training techniques into practice and contrast their resistance effects to state perturbations.
\subsection{Environments}
We implement our adversarial attacks and adversarial training on the StarCraft Multi-Agent Challenge (SMAC). StarCraft II is a commonly used multi-agent training environment widely used in MARL. We use four SMAC maps as the environments of our experiments: 2m\_vs\_1z (2 Marines vs. 1 Zealot), 3m (3 Marines vs. 3 Marines), 3s\_vs\_3z (3 Stalkers vs. 3 Zealots) and 2s3z (2 Stalkers \& 3 Zealots vs. 2 Stalkers \& 3 Zealots). More details about SMAC can be found in \cite{b20}. We consider the most extreme case in which dense attacks disturb all multiple agents at each timestep ($M=N$). We use the mainstream c-MARL algorithm QMIX and train it for robustness based on the gradient method, policy regularization, ATLA, and PA-AD, respectively.
\subsection{Adversarial Attacks and Trainings}
\textbf{Training.} In the training phase, we attack $M$ agents to increase QMIX's resilience. For the 2m\_vs\_1z, 3m, and 3s\_vs\_3z maps, we set the perturbation magnitude to 0.2, and for the 2s3z map, to 0.05. The training timestep of the vanilla QMIX models (the ones trained with clean states) is 205,000, while the training timestep of the QMIX models with robust training is 405,000. The gradient-based optimization method we choose is FGSM. In the policy regularization model, we first use the already-trained QMIX model as a pre-training model. Then we use the model to add the regular term training under FGSM interference, where $\kappa$ is initially set to 0 in the first half of training and is given a value of 0.1 in the second half of training. The hinge value $c$ is set to 10. In the ATLA and PA-AD training, we let the training object network and the adversarial network cross-train at separate intervals to improve the performance of both continuously. We used the MAPPO algorithm as the adversary during the ATLA training, with the learning rates of both actor and critic set to 1e-6 and a gradient clip value of 10. The adversarial algorithm used for PA-AD training is QMIX.

\textbf{Testing.} 
In the testing phase, we launched attacks on each of the five training models in the four maps with all the attack methods used during training except for policy regularization (which is a defense method and not an attack method). Each round of the test phase was played 32 times each, with the same perturbation magnitude as the training phase, and still attacked all the agents. Table ~\ref{tab1} shows the results of the cross-attack of these methods.

\begin{table*}[h]
\label{table1}
\caption{Cross-attacks in the Testing Phase}
\begin{center}
\begin{tabular}{|c|c|c|c|c|c|c|c|c|c|}
\hline
\textbf{Maps}&\textbf{Training Methods}&
\multicolumn{2}{|c|}{\textbf{No Attack}}&
\multicolumn{2}{|c|}{\textbf{FGSM Attack}}&
\multicolumn{2}{|c|}{\textbf{ATLA Attack}}&
\multicolumn{2}{|c|}{\textbf{PA-AD Attack}}\\
\cline{3-10} 
& &
\textbf{\textit{Win$^{\mathrm{*}}$}}& \textbf{\textit{Reward}}& 
\textbf{\textit{Win}}& \textbf{\textit{Reward}}& 
\textbf{\textit{Win}}& \textbf{\textit{Reward}}&
\textbf{\textit{Win}}& \textbf{\textit{Reward}}\\
\hline
&Vanilla QMIX                   &1.00&20.0000$\pm$0.0000&0.00&2.3656$\pm$0.4219&0.00&3.7334$\pm$0.2981&0.00&2.3602$\pm$0.3102\\
&FGSM                           &0.87&18.2787$\pm$2.9210&1.00&20.0000$\pm$0.0000&1.00&20.0000$\pm$0.0000&0.85&18.1783$\pm$2.8507\\
2m\_vs\_1z&Policy Regularization&1.00&20.0000$\pm$0.0000&0.94&19.155$\pm$2.1991&0.98&19.7966$\pm$1.1141&0.77&16.9552$\pm$3.3790\\
&ATLA                           &0.53&13.8056$\pm$1.6720&0.00&3.0287$\pm$0.6546&1.00&20.0000$\pm$0.0000&0.00&2.8701$\pm$0.1548\\
&PA-AD                          &1.00&20.0000$\pm$0.0000&1.00&20.0000$\pm$0.0000&1.00&20.0000$\pm$0.0000&0.98&19.8011$\pm$1.0896\\
\hline
&Vanilla QMIX                   &0.98&19.8065$\pm$1.0601&0.00&4.7963$\pm$0.3710&0.81&17.5908$\pm$3.0427&0.00&2.2996$\pm$0.3254\\
&FGSM                           &0.92&18.9951$\pm$2.2924&0.97&19.6129$\pm$1.4746&0.97&19.5970$\pm$1.5347&0.77&17.1357$\pm$3.1627\\
3m&Policy Regularization        &0.34&11.5961$\pm$3.0177&0.97&19.6235$\pm$1.4336&0.94&19.2311$\pm$1.9982&0.82&17.7826$\pm$2.9991\\
&ATLA                           &0.90&18.6328$\pm$2.8014&0.00&5.3283$\pm$0.5470&0.02&5.6368$\pm$1.3908&0.02&5.5272$\pm$1.3579\\
&PA-AD                          &1.00&20.0000$\pm$0.0000&0.97&19.6076$\pm$1.4955&0.90&18.8228$\pm$2.4047&0.97&19.6129$\pm$1.4746\\
\hline
&Vanilla QMIX                   &1.00&20.0000$\pm$0.0000&0.00&7.8688$\pm$1.2552&0.74&18.4149$\pm$1.5862
&0.00&5.1305$\pm$0.6753\\
&FGSM                           &1.00&20.0000$\pm$0.0000&1.00&20.0000$\pm$0.0000&1.00&20.0000$\pm$0.0000&0.84&18.8243$\pm$1.9965\\
3s\_vs\_3z&Policy Regularization&1.00&20.0000$\pm$0.0000&1.00&20.0000$\pm$0.0000&1.00&20.0000$\pm$0.0000&0.82&18.5683$\pm$2.0629\\
&ATLA                           &1.00&20.0000$\pm$0.0000&0.00&8.1845$\pm$0.7114&1.00&20.0000$\pm$0.0000&0.00&8.2365$\pm$1.0631\\
&PA-AD                          &1.00&20.0000$\pm$0.0000&0.98&19.9029$\pm$0.5543&1.00&20.0000$\pm$0.0000&0.90&19.3391$\pm$1.3766\\
\hline
&Vanilla QMIX                   &1.00&20.0000$\pm$0.0000&0.05&12.2415$\pm$1.3581&0.90&19.3830$\pm$1.3067&0.02&13.7166$\pm$1.1353\\
&FGSM                           &1.00&20.0000$\pm$0.0000&1.00&20.0000$\pm$0.0000&1.00&20.0000$\pm$0.0000&0.94&19.6231$\pm$1.0591\\
2s3z&Policy Regularization      &0.98&19.9031$\pm$0.5341&0.95&19.7577$\pm$0.8400&0.98&19.9456$\pm$0.3851&0.95&19.6771$\pm$0.9985\\
&ATLA                           &1.00&20.0000$\pm$0.0000&0.03&12.5423$\pm$1.2042&1.00&20.0000$\pm$0.0000&0.13&13.2765$\pm$1.8647\\
&PA-AD                          &1.00&20.0000$\pm$0.0000&0.95&19.7056$\pm$0.9563&1.00&20.0000$\pm$0.0000&0.97&19.9701$\pm$0.4767\\
\hline
\multicolumn{4}{l}{$^{\mathrm{*}}$Winning rate in the testing period.}
\end{tabular}
\label{tab1}
\end{center}
\end{table*}

\subsection{Results Analysis}
We evaluate the pros and cons of these four strategies to improve the robustness of MARL based on the trials mentioned earlier. There are benefits and drawbacks to each of these four strategies in terms of training difficulty and the magnitude of enhancements.
\begin{itemize}
\item $\textbf{Gradient-based adversary}$ is an easy and effective way to boost the robustness of MARL in the training period. But it does not provide the strongest attack, making the trained QMIX algorithm not robust enough to resist stronger attacks.
\item $\textbf{Policy Regularization}$ is based on the relationship between the values of the clean and attacked states by adding regular terms to the loss function. Unfortunately, this approach performs poorly when stronger interference is present, such as optimal adversary. Furthermore, it doesn't act stably in clean states.
\item $\textbf{ATLA}$ generates optimal state perturbations by training a network, which would theoretically perform better in SARL scenarios with small state spaces with a significant improvement in robustness. However, in the MARL scenario, the state adversarial network must produce numerous state perturbations for multiple agents, multiplying the network's state and action space. This is especially true in many environments where the input images are used as states, which makes training even more challenging. Gradient explosion is a problem that we ran into during the training process, and even when the training is completed, the outcomes are highly disappointing.
\item $\textbf{PA-AD}$ is similar to ATLA in idea, training a state adversarial network to attack the observed state of an agent. The distinction is that PA-AD creates the perturbation direction and employs state interference by that perturbation direction, thus reducing the action space and simplifying network training.
\end{itemize}
\section{Conclusions and Future Work}
In this paper, we migrate the robustness training from the single-agent case to the multi-agent scenario. We also describe four approaches to enhance the robustness of the widely used MARL algorithm QMIX and discuss the theoretical foundation for these methods in the multi-agent setting. Based on these, we put each of these training techniques into practice and determine how effectively they improve learning. In future work, we will optimize the adversarial networks used in the training process, explore more robust training methods that combine high efficiency and good results, and apply them to MARL algorithms other than QMIX.

\section*{Acknowledgment}

\end{document}